# Free congruence: an exploration of expanded similarity measures for time series data

**Lucas Cassiel Jacaruso[1]**
[1]University of Southern California, Los Angeles, CA 90007 USA
[1]Universität für Musik und darstellende Kunst Graz, Mondscheingasse 7, 8010, Graz, Steiermark, Austria

Corresponding Author: Lucas Cassiel Jacaruso (personal email (preferred): jacaruso33@gmail.com, institutional email: jacaruso@usc.edu)

**ABSTRACT** Time series similarity measures are highly relevant in a wide range of emerging applications including training machine learning models, classification, and predictive modeling. Standard similarity measures for time series most often involve point-to-point distance measures including Euclidean distance and Dynamic Time Warping. Such similarity measures fundamentally require the fluctuation of values in the time series being compared to follow a corresponding order or cadence for similarity to be established. Other existing approaches use local statistical tests to detect structural changes in time series. This paper is spurred by the exploration of a broader definition of similarity, namely one that takes into account the sheer numerical resemblance between sets of statistical properties for time series segments irrespectively of value labeling. Further, the presence of common pattern components between time series segments was examined even if they occur in a permuted order, which would not necessarily satisfy the criteria of more conventional point-to-point distance measures. The newly defined similarity measures were tested on time series data representing over 20 years of cooperation intent expressed in global media sentiment. Tests determined whether the newly defined similarity measures would accurately identify stronger resemblance, on average, for pairings of similar time series segments (exhibiting overall decline) than pairings of differing segments (exhibiting overall decline and overall rise). The ability to identify patterns other than the obvious overall rise or decline that can accurately relate samples is regarded as a first step towards assessing the value of the newly explored similarity measures for classification or prediction. Results were compared with those of Dynamic Time Warping on the same data for context. Surprisingly, the test for numerical resemblance between sets of statistical properties established stronger resemblance for pairings of decline years with greater statistical significance than Dynamic Time Warping on the particular data and sample size used.

## 1. INTRODUCTION

Similarity measures between datasets are pertinent to virtually every area of science and computing, with applications ranging from classification and machine learning to forecasting and beyond. To date, many studies have been conducted on similarity measures as specifically applied to time series data. Pattern sequence similarity has been of particular interest in forecasting series of economic significance, such as electricity demand and solar power output. [1], [2]. A common approach is to make use of distance measures to identify samples of a time series similar to a period of interest, and use the most similar historical samples to inform a forecast of subsequent changes. Such approaches can be generally categorized as clustering methods. Distance measures are a way of quantifying pattern similarity between time series by measuring the proximity between points they comprise. Sun et al. [3] employ point-to-point Euclidean distance in conjunction with Angle Cosine for clustering similar segments of time series data in the context of wind power forecasting, namely by using the similar samples as training data for a neural network. In another study, Bandara et al. [4] also apply distance-based clustering to the training of a neural network in the context of forecasting. Dynamic Time Warping [5], commonly abbreviated as DTW, is an applicable similarity measure when the resemblance between time series isn't temporally synced. The minimum degree of "warping" necessary to reasonably sync the series for a best match (in terms of point-to-point-distance) can serve as a similarity measure. Yu et. al [6] demonstrate the applicability of DTW alongside a neural network to forecast peak load in power demand. Studies have also investigated the use of DTW solely for time series classification purposes without the forecasting element [7, 8, 9]. As demonstrated in a review of similarity measures by Serrà et al. [10], point-to-point distance measures are commonly implemented on the features of time series data rather than the raw time series itself by using Fourier coefficients. In a separate category from the aforementioned research, some work has also been done regarding comparison of local statistical test results to detect changes in a time series. Kosiorowski et al. [11] use an adaptation of the Wilcoxon rank sum test for detecting structural changes in samples of a time series. Tang et al. [12] compare the statistical properties of time series segments classified into flow states. In their review, Zou et a. [13] detail statistical similarity between time series segments as a rationale for choosing proximity networks such as recurrent networks [14] for time series analysis. Still, the use of local statistical tests on time series samples in the specific context of similarity measures or for change detection remains less frequently studied.

Whether applying similarity measures for forecasting or classification purposes, the majority of studies approach similarity measures in terms of point-to-point distance such as Euclidean distance or DTW. Both aforementioned methods require the rise and fall of values in the time series being compared to follow a similar order or succession, even if the resemblance isn't temporally synced (as is the main application for DTW).

In order to illustrate how the notions of similarity explored in this paper are distinct from conventional similarity measures, we must first clearly define what will be called "super-sequences" and "fluctuation subsequences" going forward. A super-sequence is simply defined as a sequence of values within a temporal range of a time series delineated by a start and end, in this case a date range. It can be thought of simply as a historical segment of a time series. A fluctuation subsequence will be defined as the sequence of percent change magnitudes from value-to-value in a proper subsequence of the super-sequence. Going forward, fluctuation subsequences will be treated as attributes of super-sequences.

The first similarity test quantified the numerical resemblance between sets of descriptive statistical properties of the two super-sequences. The set of statistical properties of the super-sequences is defined as a finite set of the following discrete values: mean, standard deviation, minimum, maximum, twenty fifth percentile value, fiftieth percentile value, and seventy fifth percentile value of the super-sequence. The sheer numerical resemblance between such sets was established regardless of the "ground truth" labelling of the values. Otherwise put, the presence of roughly equal values in sets of statistical properties, even if they aren't allocated to the same property, would still establish a degree of resemblance. This is a departure from the norm in that statistical similarity usually means properties must be similar (e.g. similar mean values).

The second test was simply based on the number of fluctuation subsequences in common between the two super-sequences, even if the super-sequences do not bear resemblance in terms of point-to-point distance. In other words, a match was established if common fluctuation subsequences were at all present (even in a permuted order), which would not necessarily satisfy the criteria of point-to-point distance measures; the latter requires patterns to occur in the corresponding order. Since point-to-point distance measures serve as the conceptual underpinning of most similarity measures including those used in clustering, this is also a departure from the norm in the context of the application.

The newly explored notion of similarity as untethered to ground-truth labelling of statistical properties or pattern permutation order will be termed "free congruence". The practical significance of intersecting descriptive statistical values despite their labelling, and common pattern components despite permutation order, is a seemingly unlikely yet interesting possibility to methodically explore in the context of time series similarity. This study addresses whether free congruence would accurately suggest stronger similarity between an exemplary period of decline and other periods of decline in a time series than between the said exemplary decline period and periods of growth in the same time series. As demonstrated by related work, similarity measures can draw comparisons between time series samples such that similar samples can successfully inform predictions. If the newly explored approaches can first successfully identify similarity in a way that successfully establishes stronger resemblance between decline years (other than the obvious feature of overall decline itself), further study may be warranted for the application of these similarity measures as classification tools or even predictive indicators. Limitations of this study include the fact that predictive value of the newly explored similarity measures was not yet evaluated, nor were the parameters for classification fine-tuned with training and testing data; simply the ability to rate samples for similarity was examined as an initial step. Tests were carried out on a time series dataset representing cooperative intent expressed in the global media sentiment from 1992 to 2017. A year of markedly decreased total cooperative tone in the media was chosen and compared with other years of both decreased and increased cooperative tone for similarity within the new definitions. Results of the new method were compared with those of DTW on the same data for context.

## 2. METHODS AND RESULTS

*2.1 DATA*

Tests were carried out on a time series dataset representing global cooperative tone in the media since 1992 obtained from GDELT, or the Global Database of Events, Language and Tone [15]. GDELT aggregates data from global print, broadcast, and web news and applies natural language processing to quantify events and sentiment both globally and by region. The data were accessed via Google BigQuery [16] and queried via standard SQL. The exact GDELT eventcode [17] used to obtain the data was 03, representing 'express intent to cooperate' (not to be confused with 'engage in material cooperation' which is an available filter assigned to another eventcode). Fraction dates represent the percent of the year completed at a given time (ranging from 0 to 0.9999) and are a way of roughly standardizing the temporal distance between dates. Fraction dates do not take into account leap years or the varying length of the months. The final data used were normalized as a percent of total, namely the percent of cooperative tone instances in the media out of the total news press (comprising all event codes) for each fractiondate.

*2.2 TESTING PROCESS*

The test process for determining the degree of similarity between two segments of a time series can be broadly outlined as follows:

Firstly, a super-sequence $Y$ was defined which contained the values of the time series representing a recent year of decline. Otherwise put, $Y$ is simply a segment of the whole time series dataset defined by the start and end fractiondates of the decline year of interest. By year of decline is meant a year that exhibited a decrease in the running total of all time series values for the year as compared to the same for the values of the previous year. $Y$ was the time series segment representing cooperative tone in the media for fractiondates in 2019 which saw a 2.63% decrease in the running total of values as compared the same for 2018. For each test conducted, a separate super-sequence $X$ was also selected. $X$ represented an earlier segment of the time series defined in the same way as $Y$, namely by a by start and end date encompassing a year's worth of time series data points. The timeframe of $X$ always preceded that of $Y$. The temporal resolution and length of $X$ and $Y$ were strictly identical in each individual test. Two sets of tests were conducted. Depending on the test set, $X$ could represent a year of total decline or a year of total growth compared to the previous year (the start and end delineations of sets for each test will be detailed later). One test set was to determine the average measure of similarity between $Y$ and a succession of ten $X$ sequences each representing past periods of decline in the time series. The other test set determined average similarity between $Y$ and a succession of ten $X$ super-sequences each representing past periods of growth. As previously stated, $Y$ was always the same for every test in both test sets; only $X$ changed for each individual test. The purpose of structuring the tests as such was to establish whether the similarity measures used in this study would successfully point to a stronger resemblance on average between $Y$ and other years of decline than between $Y$ and years of growth in any statistically significant way.

For each unique pairing of $X$ and $Y$ super-sequences for every test, three measures of similarity were established. The first measure addressed the numerical resemblance between their respective statistical property sets, and the second addressed common fluctuation subsequences. Finally, DTW distance was calculated.

*2.3 NUMERICAL RESEMBLANCE BETWEEN STATISTICAL PROPERTY SETS*

For each unique pairing of $Y$ and $X$, this first test established a set of statistical properties describing $Y$ which will be referred to as $K$. Another set of the same properties was established for $X$, which will be referred to as $L$. Sets $K$ and $L$ included the standard deviation, minimum, maximum, twenty fifth percentile value, fiftieth percentile value, and seventy fifth percentile value for $Y$ and $X$, respectively. Once sets $K$ and $L$ were established, the presence of each value in $K$ was checked for in $L$, with a 0.3 neighborhood of tolerable range. In other words, any given value in $K$ was considered to be shared with set $L$ if another value within a 0.3 range above or below the original value in question was present in $L$.

The number of common values between the sets was counted regardless of whether those shared values were representative of the same statistical property; the presence of qualifying values anywhere in the respective sets sufficed. Therefore, final result was the percentage of values in $K$ also present in $L$ within a 0.3 neighborhood. More formally, this process is denoted as follows:

For the sake of example, the denotation of statistical properties will be illustrated in the context of hypothetical set $A$. Let $\overline{A}$ signify the arithmetic mean of values in the set. Let $\sigma A$ signify the standard deviation of the set. Let $min(A)$ signify the minimum of the set, and $max(A)$ signify the maximum. Let percentile values be given by

$$P = min(A) + (max(A) - min(A)) * percentile,$$

where $percentile$ is the percentile value being calculated (e.g. 25%).

Let $A_{25\%}, A_{50\%}$ and $A_{75\%}$ denote the twenty fifth, fiftieth, and seventy fifth percentile values. Let $rd_{\frac{1}{100}}$ indicate rounding to the nearest hundredth. For example, $rd_{\frac{1}{100}}(x)$ would denote $x$ rounded to the nearest hundredth.

In this light, let $K = \{\overline{Y}, \sigma Y, min(Y), max(Y), Y_{25\%}, Y_{50\%}, Y_{75\%}\}$ and $L = \{\overline{X}, \sigma X, min(X), max(X), X_{25\%}, X_{50\%}, X_{75\%}\}$

Let $K^{(0.3)}$ denote the 0.3 neighborhood for all values in $K$:

$$K^{(0.3)} = \{x : |x - a| \leq 0.3\}\ a \in K$$

Let the set of common values between sets $K^{(0.3)}$ and $L$ be denoted as

$$I = K^{(0.3)} \cap L = \{X : X \in K^{(0.3)} \text{ and } X \in L\}$$

The final result of the test is given by:

$$G = rd_{\frac{1}{100}}((length(I)/length(K)) \times 100)$$

## 2.4 FLUCTUATION SUBSEQUENCES

After quantifying the numerical resemblance between sets of statistical properties as detailed above, a second test was carried out on the same pairing of $Y$ and $X$. This measure of similarity was the number of common fluctuation subsequences between $Y$ and $X$ which occurred at least as frequently in $X$ as in $Y$. The final resulting value for this first test was the sum of the length of all such qualifying fluctuation sequences out of the total number of value-to-value percent changes in $Y$, expressed as a percent. More formally, this process is defined as follows:

Let $\leqslant$ denote a contiguous, proper subsequence which occurs at least as often in one sequence as in another. As an example for the sake of clarity, $i$, $p$, and $l$ are hypothetical sequences. $i \leqslant p \ \& \ l$ would indicate $i$ is a contiguous, proper subsequence contained in both $p$ and $l$ and that $i$ occurs in sequence $p$ at least as often as it does in sequence $l$. In keeping with convention, let $\subseteq$ signify a subset.

For each test, $(X_t)_{t \in I}$ denotes the entire indexed sequence of elements in $X$ and $(Y_t)_{t \in I}$ denotes the entire indexed sequence of elements in $Y$, where $I$ is the index set for both. Since $X$ and $Y$ are always of the same length and temporal resolution, they can be indexed by means of the same index set. Let $T = \{I_1, I_{n-1}\}$ or the index set $I$ only up until the second-to-last value. $T$ will serve as an alternative index set which will be useful for operations going forward.

To quantify common fluctuation subsequences, the sequences of value-to-value percent change magnitudes must be first determined for $X$ and $Y$.

Let $O = rd_{\frac{1}{100}}(100 \times (\frac{|X_n - X_{n+1}|}{X_n}))_n \in T$. $O$ signifies the sequence of value-to-value percent change magnitudes (in terms of absolute value) between values in $X$.

Let $A = rd_{\frac{1}{100}}(100 \times (\frac{|Y_n - Y_{n+1}|}{Y_n}))_n \in T$. $A$ signifies the sequence the same for $Y$.

$U$ will be defined as a common index set for both $O$ and $A$.

Let the set of all possible subsequences of $A$ be denoted as:

$$N = \{(A_n)_n \in J \text{ where } J \subseteq U\}$$

In other words, $N$ is a set of all possible subsequences of $A$ obtained by restricting the index set $J$ to a possible subset of $U$. The index subset $J$ will be referred to going forward.

The symbol $\prod$ will be used to denote a concatenation of values in multiple sets or sequences.

Let sequence $R$ be defined as the concatenation of values in index subsets $J$ for all sequences which comprise set $N$, given by:

$$R = \prod_{\forall i \in N} J$$

Therefore, let $C = \{(A_n)_{n \in J} \in N \text{ where } ((A_n)_{n \in J}) \leqslant O \text{ \& } A \text{ and }$ $length((A_n)_{n \in J}) >= 2\}$ *such that R is nonrepetitive*

In summary, $C$ is a set of all possible subsequences of $A$ that are also a proper subsequence of $O$, that occur at least as often in $O$ as in $A$ and that are at least two values in length, such that $R$ is a non repeating sequence. The last condition ensures only the longest possible qualifying subsequences are represented without any overlap of smaller possible sequences already contained within a larger subsequence that has been accounted for. This makes for $J$ index sets which can be concatenated into a sequence with no repeated values, hence the condition that $R$ is nonrepetitive.

Sequence $M$ will be defined as a concatenation of all sequences in set $C$.

$$M = \prod_{\forall i \in C} i$$

The final indicator will simply be the length of $M$ divided by the length of $A$, expressed as a percentage:

$$F = rd_{\frac{1}{100}}((length(M)/length(A)) \times 100)$$

where $F$ is the final result of the test.

## 2.5 DYNAMIC TIME WARPING

DTW distance was also calculated for each pairing of $Y$ and $X$. The results of the DTW algorithm were compared with the results of the newly defined similarity measures. DTW is an algorithm for quantifying optimal alignment between two time series or temporal sequences which differ in speed or cadence. Unlike pure Euclidean distance, DTW performs one-to-many and many-to-one matches between data points so that sequential matches in peaks and troughs between datasets can be identified even if they aren't exactly synced in time. DTW was chosen for comparison with the newly explored methods because it can identify similarity where pure Euclidean alone distance may not. As a general definition, DTW calculates the optimal alignment between two sequences or time series by matching each index from one sequence to an index in the other sequence (and vice versa), generating a non-linear "warping" between the series. The degree of warping required to align the sequences optimally is considered to be the cost, and this cost can be used as a similarity measure between two time series. Intuitively, the lower the cost, the more similar the sequences being compared. DTW distance is calculated via an O(nm) algorithm based on dynamic programming to contend with the computational complexity that comes with calculating every possible warping path. FastDTW [18] was implemented in Python [19] via the FastDTW library [20] to calculate DTW distance.

DTW performed on two hypothetical time series $A$ and $B$ is more formally generalized as follows in accordance with the standard definition [5]:

$$A = (a_1, a_2, ..., a_N), N \in \mathbb{N} \text{ and } B = (b_1, b_2, ..., b_M) M \in \mathbb{N}$$

It is assumed $A$ and $B$ sequences sampled at equidistant points in time. $A$ and $B$ can be considered to be feature sequences drawing values from a feature space, denoted as $\Phi$. By extension, $a_n, b_m \in \Phi \text{ for } n \in [1:N] \text{ and } m \in [1:M]$. The comparison of sequences $A, B \in \Phi$ is based on a local distance measure, also referred to as a local cost measure. The terms "cost" and "distance" in this context are interchangeable. The local cost measure is the distance function, defined as follows:

$$c : \Phi \times \Phi \to \mathbb{R} \geq 0$$

The closer the resemblance between sequences $A$ and $B$, the lower resultant value $c$. The local cost measure for all pairs of values of $A$ and $B$ is represented in a matrix $C \in \mathbb{R}^{N \times M}$, and so $C(n, m) = c(a_n, b_m)$. The alignment between $A$ and $B$ such that the cost is kept to a minimum will be the optimal alignment. The optimal alignment path can be algorithmically determined, and is a path covering the lowest cost regions of the matrix. The term alignment path is synonymous with warping path in the conventional nomenclature around DTW.

A possible $(N, M)$ warping path is a sequence $p = (p_1, \ldots, p_L)$ with $p_l = (n_l, m_l) \in [1:N] \times [1:M]\ for\ l \in [1:L]$ such that the following conditions are met:

### (i) Boundary Condition:

$p_1 = (1, 1)$ and $p_L = (N, M)$. Simply put, the warping path must have start and end points which are the first and last points of the sequences to be aligned.

### (ii) Monotonicity Condition:

The time ordering of points must be maintained via a condition of monotonicity:

$n_1 \leq n_2 \leq \ldots \leq n_L$ and $m_1 \leq m_2 \leq \ldots \leq m_L$

### (iii) Step Size Condition:

$p_{l+1} - p_l \in \{(1,1), (1,0), (0,1)\}\ for\ l \in [1:L-1]$. This ensures all index pairs in the warping path are pairwise distinct and no elements in $A$ and $B$ are left out.

The warping path $p = (p_1, \ldots, p_L)$, therefore, makes for an alignment between sequences $A$ and $B$ by associating elements of $A$ to elements of $B$.

The total cost for a warping path in the context of the cost measure $c$ is:

$$c_p(A, B) = \sum_{l=1}^{L} c(a_{n_l}, b_{m_l})$$

Out of all possible warping paths, let $p*$ denote the warping path of minimum total cost. Therefore, let the DTW distance between $A$ and $B$, or $DTW(A, B)$, be defined as simply the total cost of $p*$:

$$DTW(A, B) = c_{p*}(A, B) = min\{c_p(A, B) \text{ where } p \text{ is an } (N, M) \text{ warping path}\}$$

## 2.6 RESULTS

The test results for similarity measures between the decline year and other historical decline years (test set one) are as follows:

**Table 1: Test results for similarity measures between 2019 and other historical decline years (test set one)**

| Years Compared | | Similarity Measure Indicator | | |
|---|---|---|---|---|
| Decline Year $X$ | Decline Year $Y$ | $F$ Value (Fluctuation Subsequences) | $G$ Value (Raw Numerical Resemblance of Statistical Properties) | DTW Distance |
| 2002 | 2019 | 0.28% | 28.57% | 28.12 |
| 2004 | 2019 | 1.12% | 28.57% | 23.52 |
| 2005 | 2019 | 0% | 57.14% | 21.62 |
| 2007 | 2019 | 1.68% | 57.14% | 17.21 |
| 2008 | 2019 | 0.84% | 57.14% | 13.1 |
| 2011 | 2019 | 1.68% | 71.43% | 14.98 |
| 2012 | 2019 | 0.84% | 71.43% | 12.66 |
| 2014 | 2019 | 1.4% | 85.71% | 12.05 |
| 2016 | 2019 | 1.12% | 42.86% | 30.69 |
| 2017 | 2019 | 1.4% | 71.43% | 11.79 |

Based on the results in the above table, the mean values for each of the similarity measure indicators when comparing two decline years were as follows:

**Table 2: Mean result per similarity measure when comparing two decline years**

| Mean $F$ value | 1.036 % |
|---|---|
| Mean $G$ value | 51.742% |
| Mean DTW Distance Measure | 18.574 |

Next, the test results for similarity measures between the decline year and historical rise years (test set two) are as follows:

**Table 3: Test results for similarity measures between 2019 and historical rise years (test set two)**

| Years Compared | | Similarity Measure Indicator | | |
|---|---|---|---|---|
| Rise Year $X$ | Decline Year $Y$ | $F$ Value (Fluctuation Subsequences) | $G$ Value (Raw Numerical Resemblance of Statistical Properties) | DTW Distance |
| 1992 | 2019 | 0.84% | 14.29% | 53.04 |
| 1994 | 2019 | 0% | 14.29% | 51.74 |
| 1996 | 2019 | 0.84% | 0% | 44.69 |
| 1997 | 2019 | 0% | 14.29% | 56.36 |
| 2003 | 2019 | 0.28% | 0% | 29.20 |
| 2006 | 2019 | 1.4% | 28.57% | 21.17 |
| 2009 | 2019 | 1.4% | 42.86% | 16.9 |
| 2010 | 2019 | 3.35% | 42.86% | 17.36 |
| 2013 | 2019 | 0.84% | 57.14% | 12.6 |
| 2015 | 2019 | 1.12% | 42.86% | 14.01 |

Based on the results in the table 3, the mean values for each of the similarity measure indicators when comparing the decline year with rise years was as follows:

**Table 4: Average result per similarity measure when comparing decline and rise years**

| | |
|---|---|
| Mean $F$ value | 0.995% |
| Mean $G$ value | 25.716% |
| Mean DTW Distance Measure | 32.307 |

The test results for raw numerical resemblance between sets of statistical properties, (represented by the value $G$) were significantly higher on average in test set one than in test set two at 51.742% vs 25.716%, respectively (two sample Welch t-test, *p* = 0.01).

The test results for common fluctuation subsequences, (represented by the value $F$) were not significantly higher on average in test set one than in test set two at 1.036% vs 0.995%, respectively (Wilcoxon rank sum test, $p = 0.4$). Note, a nonparametric test was chosen in this case since results were not normally distributed.

The test results for DTW distance were significantly lower on average in test set one than in test set two at 18.574 vs 32.307, respectively (two sample Welch t-test, $p = 0.04$).

## 3. DISCUSSION

The mean degree of resemblance indicated by $G$ value tests for pairings of decline years (in test set one) was higher than for pairings of rise and decline years (in test set two) at 51.742% vs 25.716%, respectively. DTW distance was lower on average when comparing decline years at 18.574 vs 32.407, respectively (the lower the value, the more similar). The difference in average $G$ value test results between test sets was more statistically significant than the difference in average DTW distance results between test sets, at $p = 0.01$ vs $p = 0.04$, respectively. $F$ values, representing common fluctuation subsequences, were not significantly higher on average for pairings of decline years. The results suggest some degree of validity to the newly defined notions of similarity worthy of further study specifically in terms of numerical resemblance between sets of statistical properties. Upon considering the results, the glaring question became whether the numerical overlaps between statistical property sets were actually attributable to values allocated to the same properties or different properties. Across all $G$ value tests in all test sets, only 56.9% of overlapping values between sets of statistical properties were found to represent the same property. This substantiates one aspect of the notion of free congruence; numerical resemblance alone is valuable without the criterion that overlapping values must necessarily represent the same property. This is a surprising result which is, at the very least, worthy of additional investigation. Limitations of this study call for the phenomenon to be examined on larger sample sizes and for further inquiry into possible explanations for the observation.

## 4. CONCLUSION

This paper explores a broadened definition of similarity measures in the context of time series data. The sheer numerical resemblance between sets of statistical properties of time series segments is explored as a similarity measure. Shared fluctuation patterns between time series segments are also explored for establishing similarity. These measures of similarity are defined as free congruence. The empirical results suggest some degree validity to the notion of testing numerical resemblance of statistical property sets for time series segments in the context of the application, even if intersecting values (within a 0.3 neighborhood) aren't allocated to the same statistical property. Surprisingly, this latter method identified stronger resemblance on average between similar time series samples versus differing samples, with greater statistical significance than Dynamic Time Warping on the data and sample size used. Further study is called for to test the validity of free congruence (specifically in the context of numerical resemblance between statistical property sets) on larger time series samples and different datasets. The possibility of using the newly explored similarity measures to generate forecasts is also of interest. Free congruence may be a significant phenomenon that can complement existing similarity measures such as DTW in the relevant range of applications.

## 5. ABBREVIATIONS

*5.1 DATA*

GDELT: The Global Database of Events, Language, and Tone

*5.2 METHODS / ALGORITHMS*

DTW: Dynamic Time Warping

## 6. REFERENCES

[1] Pérez-Chacón, R., Asencio-Cortés, G., Martínez-Álvarez, F., & Troncoso, A: Big data time series forecasting based on pattern sequence similarity and its application to the electricity demand. *Information Sciences* 2020, *540*, 160-174.

[2] Wang, Z., Koprinska, I., & Rana, M: Solar power forecasting using pattern sequences. In *International Conference on Artificial Neural Networks* 2017 (pp. 486-494). Springer, Cham.

## 7. DECLARATIONS

*8.1 FUNDING*

This research did not receive any specific grant from funding agencies in the public, commercial, or not-for-profit sectors.

*8.2 AUTHOR'S CONTRIBUTIONS*

LJ developed the new similarity measures put forth and explored in this paper, obtained and prepared the data, and authored all aspects of the manuscript.

*8.3 COMPETING INTERESTS*

The author declares that they have no competing interests.

*8.4 ACKNOWLEDGEMENTS*

This research would not have been possible without the ongoing support of Mimi and Don Jacaruso and Luciana DeClemente, family of the author.

*8.5 AVAILABILITY OF DATA AND MATERIALS*

Specific datasets related to the research presented in this paper can be obtained as detailed in the work, or can be obtained from the corresponding author upon reasonable request.